\documentclass[conference]{IEEEtran}
\IEEEoverridecommandlockouts
\usepackage{cite}
\usepackage{amsmath,amssymb,amsfonts}
\usepackage{algorithmic}
\usepackage{graphicx}
\usepackage{textcomp}
\usepackage{xcolor}
\usepackage{adjustbox}
\usepackage{resizegather}
\usepackage{subfigure}
\usepackage{multicol}
\usepackage{multirow}
\usepackage{flushend}
\usepackage{makecell}
\usepackage{bm}
\def\BibTeX{{\rm B\kern-.05em{\sc i\kern-.025em b}\kern-.08em
    T\kern-.1667em\lower.7ex\hbox{E}\kern-.125emX}}
\usepackage{algorithm2e}

\begin{document}

\title{BSSAD: Towards A Novel Bayesian State-Space Approach for Anomaly Detection in Multivariate Time Series \\}

\author{\IEEEauthorblockN{Usman Anjum}
\IEEEauthorblockA{\textit{University of Cincinnati} \\
Cincinnati, OH, USA \\
anjumun@ucmail.uc.edu}
\and
\IEEEauthorblockN{Samuel Lin}
\IEEEauthorblockA{\textit{University of Arkansas}  \\
Fayetteville, Arkansas, USA \\
smlin@uark.edu}
\and
\IEEEauthorblockN{Jusin Zhan}
\IEEEauthorblockA{\textit{University of Cincinnati} \\
Cincinnati, OH, USA \\
zhanjt@ucmail.uc.edu}
}

\maketitle

\begin{abstract}

Detecting anomalies in multivariate time series (MTS) data plays an important role in many domains. The abnormal values could indicate events, medical abnormalities, cyber-attacks, or faulty devices which if left undetected could lead to significant loss of resources, capital, or human lives. In this paper, we propose a novel and innovative approach to anomaly detection called \textit{Bayesian State-Space Anomaly Detection (\texttt{BSSAD})}. The \texttt{BSSAD} consists of two modules: the neural network module and the Bayesian state-space module. The design of our approach combines the strength of Bayesian state-space algorithms in predicting the next state and the effectiveness of recurrent neural networks and autoencoders at understanding the relationship between the data to achieve high accuracy in detecting anomalies. The modular design of our approach allows flexibility in implementation with the option of changing the parameters of the Bayesian state-space models or swapping neural network algorithms to achieve different levels of performance. In particular, we focus on using Bayesian state-space models of particle filters and ensemble Kalman filters. We conducted extensive experiments on five different datasets. The experimental results show the superior performance of our model over baselines, achieving an F1-score greater than 0.95. In addition, we also propose using a metric called Matthew Correlation Coefficient (MCC) to obtain more comprehensive information about the accuracy of anomaly detection. 
\end{abstract}

\begin{IEEEkeywords}
Bayesian state-space models, Particle filters, Ensemble Kalman filters, Anomaly detection
\end{IEEEkeywords}

\section{Introduction} \label{intro}

Anomaly detection in multivariate time series (MTS), also referred to as outlier or novelty detection, is the task of detecting data that deviates significantly from expected behavior. A multivariate time series consists of more than one time-dependent variable. Each variable depends not only on its past values but also has a dependency on other variables. In the medical domain or cyber-physical system, these variables could be sensors measurement like temperature, pressure, etc. Anomaly detection has applications in many domains. For example, in event detection \cite{anjum2021tbam}, in the financial domain to identify fraud, in cyber-security to identify cyber-attacks, for identifying medical risks, or for fault detection in cyber-physical systems \cite{pang2021deep}. 

Detecting anomalous behavior in multivariate time series has been an active area of research for a long time \cite{chandola2009anomaly, pang2021anomaly}. Most of the previous works have focused on presenting multivariate time series as a collection of single time dependant variables without focusing on the dependencies between the variables \cite{liu2008isolation}, \cite{audibert2020usad}. These works have been classified as residual-based anomaly detection \cite{feng2021time}. In residual-based anomaly detection, recurrent neural networks or autoencoders are usually implemented to predict future values \cite{kwon2019survey}. Then the predicted and the actual values are compared to calculate a residual error. Any error that exceeds a threshold value greater than the actual value is classified as an anomaly. Recently, there has been a focus on designing methods that consider multivariate time series as a whole by finding relationships along the temporal domain and between the features. The results from these methods have been highly promising \cite{feng2021time}, \cite{li2021multivariate}. These methods have been referred to as density-based algorithms and have been able to achieve higher accuracy compared to residual-based algorithms in detecting anomalies. 

Density-based algorithms detect anomalies in multivariate time series data using Bayesian Theory. In Bayesian theory, the next state of a system is estimated by constructing the posterior probability density function of the data from a prior distribution. It is assumed that normal data is sampled from a posterior density function different from the one that anomalous data is sampled from. Hence, any data point that does not follow the same distribution or deviates significantly from the distribution followed by the normal data will be flagged as an anomaly. The most common state estimation techniques include Kalman filters and its variants \cite{labbe2014kalman}, Sequential Monte Carlo (SMC) and Markov Chain Monte Carlo (MCMC) methods \cite{carlin2008bayesian, chopin2020introduction}.

Implementing Bayesian state-space algorithms for anomaly detection in multivariate time series has significant obstacles. One obstacle is that to properly estimate the state, key knowledge about the mathematical relationship between the features (sensors) and their values along the time domain is required. Most of the previous works on state estimation have been built using physics equations \cite{hommels2009comparison}. But that may not always be the case and definite mathematical models between the variables may not be available. Another key obstacle is that there is no information about the posterior distribution.

To deal with these obstacles, we propose a novel and innovative density-based anomaly detection algorithm called \textit{Bayesian State-Space Anomaly Detection (\texttt{BSSAD})}. Our method consists of two main modules: the neural network module (NNM) and the Bayesian state-space module (BSSM). The NNM is based on recurrent neural networks and autoencoders. The BSSM is implemented using Bayesian state-space estimation techniques of ensemble Kalman filters (EnKF) and particle filters (PF).

Our method is motivated by the density-based algorithm presented in \cite{feng2021time}. To develop mathematical relationships between the variables, we use a combination of recurrent neural networks, more specifically long short-term memory (LSTM) \cite{hochreiter1997long}, and autoencoders (AE). LSTM and AE have been widely used for anomaly detection \cite{park2018multimodal, said2020network, nguyen2021forecasting}. The Bayesian state-space techniques estimate the posterior distribution using a Monte Carlo (MC) method. In MC methods, the posterior distribution is represented as a set of random samples from a known distribution. As the number of samples increases, the true posterior probability density function (pdf) can be estimated. Both EnKF and PF are types of MC methods. 

In contrast to the algorithm in \cite{feng2021time}, which implemented a variant of the Kalman filter called an unscented Kalman filter (UKF) \cite{wan2001unscented, julier2004unscented} to detect anomalies, our work implements ensemble Kalman filters (EnKF) and particle filters (PF) for constructing the posterior distribution and detecting the anomalies. UKF, EnKF, and PF can efficiently handle continuous, multivariate, and non-linear systems. PF have an added advantage of being able to handle multimodal, extremely non-Gaussian noise, and occlusions problems \cite{labbe2014kalman}. Previous research has shown that both EnKF \cite{hommels2009comparison, muhith2022estimation} and PF \cite{arulampalam2002tutorial} have been shown to have better accuracy in estimating the next state when compared to UKF. 

Once the posterior distribution has been obtained, we calculate the Mahalanobis distance to find the distance between the current state and the predicted distribution. Any value whose distance exceeds a threshold is considered an anomaly. 

The contribution of our work is as follows:

\textbf{Formulation \& Algorithm:} We propose the novel \textit{Bayesian State-Space Anomaly Detection (\texttt{BSSAD})} \footnote{https://github.com/BSSAD2022/BSSAD} model that utilizes state-space algorithms and neural networks to detect anomalies. We implement our model using Ensemble Kalman filters (EnKF) and particle filters (PF). To the best of our knowledge, no other works have focused on using EnKF and PF for anomaly detection in multivariate times series data.

\textbf{Accuracy:} We applied our model to multiple data sets. The results of our analysis showed that our model can achieve a F1-score. In addition, we propose to use the Matthews correlation coefficient (MCC) to measure the accuracy of anomaly detection as an alternative to the F1-score. MCC has been considered as a better metric than F1-score in binary classification tasks  \cite{chicco2020advantages}. 

\textbf{Generality:} The modular design of our model allows it to be highly flexible. Different parameters can be changed and algorithms can be easily swapped out so that our method can be applied to different data types and across multiple domains.

The paper is organized as follows: In Section \ref{lit} we provide a review of related literature. In Section \ref{prelim}, we provide an overview of key concepts that are necessary for building our model. In Section \ref{method}, we present our model. In Section \ref{exp} we present our experiments and observations and finally, in Section \ref{conc} we conclude.

\section{Literature Review} \label{lit}

There are many different anomaly detection algorithms available in the literature. Typically, anomaly detection algorithms are classified as supervised or unsupervised. Some researchers have proposed other classification techniques for anomaly detection methods. In \cite{feng2021time}, anomaly detection algorithms were classified as residual-error based and density based. Residual-error based methods predict the next value and find the difference between the predicted and actual values. Density-based methods find the distribution of the data and data is flagged as an anomaly if it is not part of the distribution. 

In \cite{liu2022time}, the anomaly detection algorithms were classified as statistics-based, classification-based, vocabulary-based, and reconstruction-based. Statistics-based methods are similar in definition to density-based algorithms. The data is assumed to follow a normal distribution and a variable is classified as normal if it is within the $3\sigma$ range. Classification is usually done using labeled data with high accuracy. However, labeled data is hard to obtain, thus the use of unlabeled methods is more common. Vocabulary-based methods aim to find patterns in time-series data and data deviating from these patterns is classified as an anomaly. Finally, reconstruction-based methods are similar to residual-error based methods. 

Most work found in the literature aims at using reconstruction or residual-error based methods. One of the most common methods that have served as a benchmark for many other methods is the Isolation Forest \cite{liu2008isolation}. The isolation forest is based on a tree-based architecture to identify anomalies but it is not made for multivariate time-series data. 

Neural networks, more specifically recurrent neural networks (RNN) and autoencoders or a combination of both, have recently become very popular as a method to detect anomalies. An extensive review of the application of deep learning techniques in anomaly detection can be found in \cite{pang2021deep, chandola2009anomaly}, and \cite{kwon2019survey}. Recently, transformers have also been proposed for detecting anomalies \cite{xu2022anomaly}. They renovate transformers for anomaly detection by adding an anomaly-attention mechanism.

Most works have focused on implementing long-short term memory (LSTM), a type of RNN, for reconstruction \cite{goh2017anomaly, hundman2018detecting}. One example of using autoencoders for anomaly detection was USAD \cite{audibert2020usad, ng2011sparse, gao2022tsmae}. Methods combining both LSTM and autoencoders to detect anomalies include \cite{park2018multimodal, ikeda2018estimation, nguyen2021forecasting, said2020network} and \cite{malhotra2016lstm}. Generative adversarial networks (GAN) have also been proposed for anomaly detection \cite{du2021gan, zhou2019beatgan}. All these methods are classified as residual-error or reconstruction-based anomaly detection.

Algorithms that can be classified as density or statistical-based anomaly detection include \cite{zhang2019deep} and Omnianomaly \cite{su2019robust}. These methods use generative models to predict the distribution of the data. 

Bayesian state-space methods are also examples of density-based algorithms. Even though these methods have mainly been applied for next state estimation \cite{arulampalam2002tutorial, labbe2014kalman, haykin2004kalman}, there have been some works that have applied Bayesian state-space estimation for anomaly detection. However, most of the work has been restricted to fault detection \cite{brown2009particle} or anomaly detection in videos \cite{gao2019particle} and \cite{tariq2021anomaly}. Ensemble Kalman filters have not been used for anomaly detection and have only been implemented for state estimation in weather or medical domains \cite{muhith2022estimation}. In \cite{li2021multivariate}, hierarchical variational autoencoders combined with Markov Chain Monte Carlo (MCMC) is used to learn temporal and inter-sensor correlation. A data point is flagged as an anomaly if it does not follow the temporal or inter-sensor correlation. A relevant work that has implemented Bayesian state-space algorithms was called Neural System Identification and Bayesian Filtering \cite{feng2021time}. In \cite{feng2021time} the AE and LSTM methods are improved further by adding a unscented Kalman Transform (UKF). UKF is able to learn the temporal and inter-sensor relationship and is able to approximate the distribution of the next state. 

\section{Preliminaries} \label{prelim}

\begin{figure}
    \centering
    \includegraphics[width=0.35\textwidth]{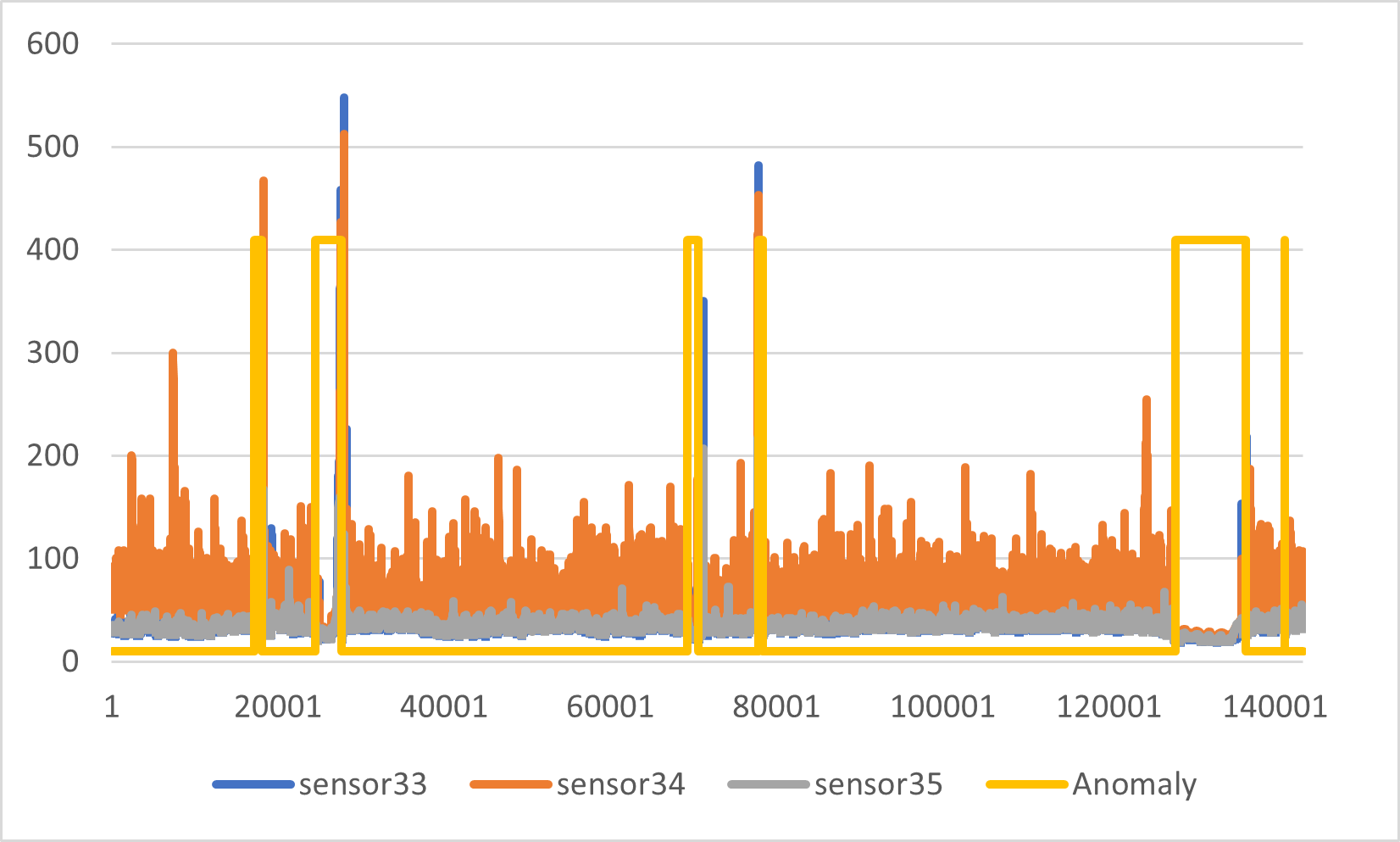}
    \caption{Anomalies in sensor data}
    \label{fig:anomalies}
\end{figure}

Figure \ref{fig:anomalies} shows how anomalies would manifest in real-world data. The anomalies would represent deviations from past patterns not only along the time-domain but also between different sensor values. In this section, we describe the Bayesian state-space model which forms the basis for our BSSM and can be used to find the relationships between the past values of the features and between different features. In particular, we present the theory behind EnKF and PF which we implement in our model.

In a Standard multivariate times series data can be represented as  $X = (\textbf{x}_0,...,\textbf{x}_T) , \textbf{x}\epsilon\mathbb{R}^{M}$where $\textbf{x}$ are $M$ features (also referred to as sensors, or variables), that can produce continuous or discrete values (when $x$ is discrete, it is referred to as \textit{control input} in the literature) and $T$ is the time interval over which the sensor values were obtained. A state-space evolution is defined by two equations:

\begin{align} \label{eq:1}
    \textbf{z}_t = F_t(\textbf{z}_{t-1}, \textbf{q}_{t-1})
\end{align}

\begin{align} \label{eq:2}
    \textbf{x}_t = H_t(\textbf{z}_t, \textbf{r}_t)
\end{align}

Equation \eqref{eq:1} is the state transition equation \cite{labbe2014kalman}. In the equation, $\textbf{z}_t$ and $\textbf{z}_{t-1}$ are the hidden state variables at time $t$ and $t-1$ respectively (e.g. actual value of a sensor), $\textbf{q}_{t-1}$ is an i.i.d. process noise, and $F_t$ is the state transition function that maps the previous state ($\textbf{z}_{t-1}$) to the current state ($\textbf{z}_t$). 

Equation \eqref{eq:2} is the measurement equation \cite{labbe2014kalman}. In the equation $\textbf{x}_t$ is the observed measurement at time $t$, $\textbf{z}_t$ is the hidden measurement (state) of the sensor, $\textbf{r}_t$ is the i.i.d. measurement noise and $H_t$ is the measurement function that maps the hidden states ($\textbf{z}_t$) to the observed measurements ($\textbf{z}_t$).

The objective is to estimate the state $\textbf{z}_t$ given the state $\textbf{x}_{1:t} = {\textbf{x}_i, i = 1, ...., t}$ up to time $T$. From a Bayesian perspective, the objective is to construct the pdf $p(\textbf{z}_t|\textbf{x}_{1:t})$. The pdf is constructed recursively in two stages: prediction and update \cite{arulampalam2002tutorial} represented by \eqref{eq:1} and \eqref{eq:2}, respectively. It is assumed that the initial state is the pdf $p(\textbf{z}_0|\textbf{x}_0) \equiv p(\textbf{z}_0)$. In the prediction stage, the system model in \eqref{eq:1} is used to obtain the prior pdf of the state at time $t$. Assuming that the prior pdf is given as $p(\textbf{z}_{t-1}|\textbf{x}_{1:t-1})$, then the prediction equation is:

\begin{align} \label{eq:Predict}
    p(\textbf{z}_t|\textbf{x}_{1:t-1}) = \int p(\textbf{z}_t|\textbf{z}_{t-1}) p(\textbf{z}_{t-1}|\textbf{x}_{1:t-1}) d\textbf{z}_{t-1}
\end{align}

It should be noted that $p(\textbf{z}_t|\textbf{z}_{t-1})$ is equivalent to the state transition equation from \eqref{eq:1}.

In the update stage, $\textbf{z}_t$ is used to modify the prior pdf to obtain the required posterior pdf. The update stage is then defined as:
\begin{equation}
    \begin{aligned} \label{eq:Update}
        p(\textbf{z}_t|\textbf{x}_{1:t}) = \frac{p(\textbf{x}_t|\textbf{z}_t)p(\textbf{z}_t|\textbf{x}_{1:t-1})}{p(\textbf{x}_t|\textbf{x}_{1:t-1})}
    \end{aligned}
\end{equation}

where $p(\textbf{x}_t|\textbf{x}_{1:t-1}) = \int p(\textbf{x}_t|\textbf{z}_t) p(\textbf{z}_t|\textbf{x}_{1:t-1}) d\textbf{z}_t$. In the above equation $p(\textbf{x}_t|\textbf{z}_t)$ is defined from the measurement model in \eqref{eq:2}.

The solution to these equations is obtained recursively but cannot be determined analytically. The optimum solution can only be obtained in certain cases. For example, the Kalman filter assumes that the posterior distribution and the noise are Gaussian  and $F_t$ and $H_t$ are linear. However, in most cases, these assumptions are not true. In such cases, only sub-optimal solutions exist. Extended Kalman filter (EKF), Unscented Kalman filter (UKF) \cite{wan2001unscented, julier2004unscented}, Ensemble Kalman filter (EnKF) \cite{crassidis2004optimal, mackenzie2003ensemble} and Particle filters (PF) \cite{arulampalam2002tutorial} are examples that provide approximate solution to the posterior pdf \cite{arulampalam2002tutorial, labbe2014kalman, speekenbrink2016tutorial}. In sub-optimal methods, the objective is to generate samples from a known prior distribution, propagate these samples using \eqref{eq:1} and \eqref{eq:2} and obtain the posterior distribution from the result. These samples are referred to as \textit{sigma points} in UKF and EnKF and \textit{particles} in PF. In the following sections, we will describe in detail how the posterior distribution is estimated using EnKF and PF.

\subsection{Ensemble Kalman Filter} \label{EnKF}

 EnKF are non-linear approximations to the state estimation problem. The EnKF is very similar to UKF \cite{labbe2014kalman} but different in some ways. In UKF, a set of weighted sigma points are generated deterministically and are passed through the state transition and measurement functions. The mean and covariance of the sigma points are an approximate measure of the posterior pdf's state mean and covariance. However, EnKF relies on Monte Carlo simulation to generate a large number of sigma points from a known distribution. The generated sigma points are then propagated using the state transition and measurement function in a similar way to UKF. Another difference is that unlike UKF, where the sigma points are generated at each iteration, in EnKF the sigma points are randomly generated based on a distribution only in the initial stage, and are then propagated at each iteration.  

The first step in designing the EnKF is the initialization stage. In the initial stage $N$ sigma points ($\bm{\chi}$) are generated using a known mean and variance.
Like any Bayesian state-space algorithm, there is a predict stage and an update stage. The implementation of the EnKF varies in different literature \cite{labbe2014kalman}. The predict stage consists of passing the $N$ sigma points ($\bm{\chi}$) through state transition function ($F_t$) to obtain $\bm{\chi}_f$. The predict stage is implemented using the following equation:

\begin{equation} \label{eq:EnKF Predict}
    \begin{aligned}
        \bm{\chi}_f = F_t(\bm{\chi}_{t-1}) + N_q \\
    \end{aligned}
\end{equation}

$N_q$ is the state transition noise matrix.

As mentioned in Section \ref{prelim}, the purpose of the update stage is to update the sigma points and obtain the posterior mean and covariance. The posterior mean and covariance are the estimates for the posterior distribution. The updated sigma points and the posterior mean and covariance become the prior state, mean and covariance for the next iteration. There are multiple ways of implementing the update state and there is no clear consensus on which way to use \cite{labbe2014kalman}. We propose to implement the update state using the following set of equations, which gave us the best performance:

 \begin{equation}  \label{eq:EnKF Update}
    \begin{aligned}
        \bm{\chi}_h &= H_t(\bm{\chi}_f) + N_r \\
        \bm{\mu}_z &= \frac{1}{N} \sum_{1}^{N} \bm{\chi}_h \\
        P_{zz} &= \frac{1}{N} \sum_{1}^{N} [\bm{\chi}_h - \bm{\mu}_z] [\bm{\chi}_h - \bm{\mu}_z]^T \\
        P_{xz} &= \frac{1}{N-1} \sum_{1}^{N} [\bm{\chi}_f - \bm{\mu}_{t-1}] [\bm{\chi}_h - \bm{\mu}_z]^T \\
        \mathcal{K} &= P_{zz} P_{xz}^{-1} \\
        \bm{\chi}_t &= \bm{\chi}_f + \mathcal{K}[x_t - \bm{\chi}_h + \bm{e_r}] \\
        \bm{\hat{\mu}}_t &= \frac{1}{N} \sum_{1}^{N} \bm{\chi_t} \\
        \hat{P}_t &= P_{t-1} - \mathcal{K}P_{zz}\mathcal{K}^T \\
    \end{aligned}
\end{equation}

The sigma points are first passed through the measurement function to obtain the hidden state $\bm{\chi}_h$. Using $\bm{\chi}_h$ the mean $\bm{\mu}_z$ is calculated. Next, we calculate $P_{zz}$ and $P_{xz}$ which are the auto-correlation between $\bm{\chi}_h$ and the cross-correlation between $\bm{\chi}_h$ and $\bm{\chi}_f$, respectively. $P_{zz}$ and $P_{xz}$ are used to calculate the Kalman Gain, $\mathcal{K}$. In our equations, we chose to change the method of calculating $P_{xz}$ and use $\bm{\mu}_{t-1}$ instead of $\textbf{x}_{t-1}$ \cite{labbe2014kalman}. The $\mathcal{K}$ is a key parameter in the EnKF framework. It is the weight given to the current state estimate and the actual state. Its purpose is to automatically tune the filter to estimate the next state as accurately as possible. The sigma points ($\bm{\chi}_t$) are updated using the Kalman gain ($\mathcal{K}$) which serves as input sigma points for the next stage. $N_r$ is the measurement noise matrix and $\bm{e_r}$ is perturbation added to the sigma points. Finally, using the updated sigma points $\bm{\chi}_t$, the posterior pdf can be estimated as state mean $\bm{\hat{\mu}}_t$ and covariance $\hat{P}_t$. 

\subsection{Particle Filter} \label{PF}

Particle filters belong to a class of Monte Carlo (MC) methods called sequential MC (SMC) filters \cite{arulampalam2002tutorial}. They are based on the sequential importance sampling (SIS) algorithm. Other SMC algorithms include bootstrap filtering, condensation algorithm, interacting particle approximations, and survival of the fittest \cite{arulampalam2002tutorial}. Like EnKF, SMC algorithms also generate sigma points (which are called \textit{particles}) sampled from a known probability distribution. However, unlike EnKF each particle is assigned a weight. The posterior distribution can then be estimated from the weighted particles. As the number of particles reaches infinity, a true representation of posterior pdf can be obtained.

A generic SIS particle filter consists of the following steps:
\begin{itemize}
    \item Generate samples called \textit{particles} from a known distribution. 
    \item Assign each particle a weight
    \item Resample the particles based on the updated weights.
    \item Calculate the state and measurement mean and covariance using the particles and their weights.
\end{itemize}

The posterior distribution is estimated using the equation:

\begin{equation} \label{eq:PF update}
    p(\textbf{z}_t|\textbf{x}_{1:t}) \approx \sum_{i=1}^{N_s} \omega_t^i\delta(\textbf{z}_t - x_t^i)
\end{equation}
where
\begin{equation} \label{eq:weights}
    \omega_t^i \propto \omega_{t-1}^i \frac{p(\textbf{x}_t|x_t^i)p(x_t^i|x_{t-1}^i)}{q(x_t^i|x_{t-1}^i, \textbf{x}_t)}
\end{equation}

Here, $\delta(\centerdot)$ is an indicator function, $x_t^i$ are the particles that are sampled from a known distribution $q(\centerdot)$ (also called importance density). $N_s$ are the number of particles sampled from $q(\centerdot)$. $\textbf{x}_t$ is the input data, $\textbf{z}_t$ is the hidden state and $w_i^t$ is the $ith$ weight at time $t$ and $\sum_i \omega_t^i = 1 $. The weights are chosen using the principal of importance sampling and the details of the derivation of the equations can be found in \cite{arulampalam2002tutorial, speekenbrink2016tutorial}.

A common problem with particle filters is \textit{particle degeneracy}. Particle degeneracy happens when after a few iterations, all but one particle will have negligible weight. Once that happens the filter will be unable to estimate the next state. A measure for particle degeneracy is the effective sample size, $N_{eff}$. However, $N_{eff}$ cannot be measured directly and only an estimate can be calculated \cite{arulampalam2002tutorial}. The equation for $N_{eff}$ is:

\begin{equation} \label{eq:Neff}
    \hat{N}_{eff} = \frac{1}{\sum_{i=1}^{N_s}(w_t^i)^2}
\end{equation}

There are two main ways of reducing particle degeneracy. One is through good choice of importance density, and the other is through resampling \cite{arulampalam2002tutorial}. The importance density can be chosen using another state-space estimation method like Kalman filters, unscented transform, or using a Gaussian estimate. For our model, we consider using a Gaussian estimate for importance density.

The second method of resampling may be computationally expensive. Hence, a common technique is to resample whenever a significant degeneracy is observed, i.e. when $N_{eff}$ falls below a threshold $N_T$. The concept of resampling is to remove all particles with small weights and focus more on particles with large weights. Some of the resampling methods found in the literature are multinomial, stratified, residual, and systematic resampling \cite{labbe2014kalman}. In this paper, we use systemic resampling schemes as it is the most popular of the resampling algorithms and performs the best amongst all the other resampling methods \cite{arulampalam2002tutorial, labbe2014kalman}.

A variation on the SIS particle filter is the sampling importance resampling (SIR) filter. The SIR filter is an MC method that can be applied to recursive Bayesian state-space problems. The SIR filter incorporates the state transition and measurement functions to propagate and update the particles. The advantage of the SIR filter is that the importance weights are easy to calculate and importance density can be easily sampled. In an SIR filter, the importance density $q(\textbf{z}_t|\textbf{x}_{t-1}^i, \textbf{x}_{1:k}) = p(\textbf{z}_t|x_{t-1}^i)$ and the particles are resampled at each step. Consequently, the weights are given by $w_t^i \propto w_{t-1}^i p(\textbf{x}_t|x_t^i)$ where $w_{t-1}^i = \frac{1}{N_s}$. 

\section{Methodology} \label{method}

\begin{figure}
    \centering
    \includegraphics[width=0.35\textwidth]{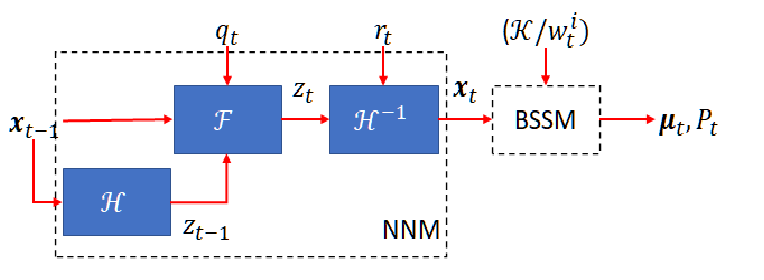}
    \caption{BSSAD Model}
    \label{fig:bssad}
\end{figure}

The overall methodology can be found in Figure \ref{fig:bssad}. Our method consists of two main modules: the neural network module (NNM) and the Bayesian state-space module (BSSM). The basis for BSSM is to use state-space methods to predict the mean and covariance of the distribution of the next state. If the sensor value deviates significantly from the predicted distribution, then it is an anomaly. The first step in the anomaly detection model is to design the state-space model. The focus of this paper is to use EnKF and SIR-PF to estimate the posterior distribution by estimating the state mean ($\bm{\hat{\mu}}_t$) and the state covariance matrix ($\bm{\hat{P}}_t$). In order to design EnKF and PF, there are four main parameters that must be determined:

\begin{itemize}
    \item state transition equation $F$ that maps the previous states to the current state
    \item process noise $q$
    \item measurement equation $H$ that maps the state to the hidden (measurement) state
    \item measurement noise $r$
\end{itemize}

Since, there are no predefined set of mathematical equations, in this paper, we follow the model presented in \cite{feng2021time} to estimate $F, q, H,$ and $r$. These parameters were estimated using recurrent neural networks. Specifically, $F$ is estimated using long short-term memory (LSTM) \cite{hochreiter1997long}. The measurement equation $H$ is estimated from an autoencoder which maps the input to a lower latent dimension. Both LSTM and autoencoders by themselves have been used for anomaly detection \cite{pang2021anomaly, kwon2019survey}. By adding a spate space algorithm that leverages autoencoders and LSTMs, the accuracy of anomaly detection can be significantly improved.

In the upcoming sections, we will describe the implementation details of our modules.

\subsection{Neural Network Module (NNM)} \label{NN}

The NNM consists of two subnets which are referred to as $\mathcal{F}$ and $\mathcal{H}$. Each of these subnets have their own parameters. The subnet $\mathcal{F}$ takes in state variables to provide times-series estimates for the state transitions. It combines the input from $\textbf{z}_{t-1}$ and $\textbf{x}_{t-\tau:t-1}$, where $\tau$ is the window size, to obtain the next state estimate $\textbf{z}_t$. It is defined as $\mathcal{F}: (\textbf{z}_{t-1}, \textbf{x}_{t-\tau:t-1})\epsilon \mathbb{R}^{M'} \rightarrow \textbf{z}_t \epsilon \mathbb{R}^{M'}$. In the autoencoder, the encoder $\mathcal{H}$ is defined as $\mathcal{H}: \textbf{x}_t \epsilon \mathbb{R}^{\tau \times M} \rightarrow \textbf{z}_t \epsilon \mathbb{R}^{\tau \times M'}$ where $M$ are the number of sensors that are mapped onto a hidden dimension $M'$. This is equivalent to the measurement function. Consequently, the decoder maps the hidden state $\textbf{z}_t$ back to the state variable $\textbf{x}_t$ and is defined as $\mathcal{H}^{-1}: \textbf{z}_t \epsilon \mathbb{R}^{W \times M'} \rightarrow \hat{\textbf{x}}_t \epsilon \mathbb{R}^{W \times M}$.

The neural network is trained using data with normal values, $X^{(n)}$. The normal data is first divided into training($X^{(n)}_{train}$) and validation set ($X^{(n)}_{val}$). $X^{(n)}_{train}$ is used to train the NNM. Following \cite{feng2021time}, the loss function of the neural network is defined as:

\begin{equation} \label{eq:loss}
    \mathcal{L} = \sum_{t=\tau}^{T} \alpha_1 ||\textbf{x}_{t-1} - \hat{\textbf{x}}_{t-1}||_2^2 + \alpha_2 ||\textbf{x}_{t} - \hat{\textbf{x}}_{t}||_2^2 + \alpha_3 ||\textbf{z}_{t} - \textbf{z}_{t-1}||_2^2
\end{equation}

where $\hat{\textbf{x}}_{t}$ and $\hat{\textbf{x}}_{t-1}$ are the reconstruction data and $\textbf{z}_{t}$ and $\textbf{z}_{t-1}$ are the hidden states for $t$ and $t-1$ respectively. In the equation, the first two terms are the reconstruction and prediction errors for sensor values. The third factor is for smoothing the consecutive hidden states and ensuring that the hidden states are close to each other. The model is trained using the ADAM \cite{kingma2014adam}. 

In addition, process and measurement noise are calculated from the neural network. The noise is calculated using  $X^{(n)}_{val}$. The process noise $q$ is found using the following equation:

\begin{equation} \label{eq:pn}
    \textbf{q}_t = \mathcal{H}(\textbf{x}_t) - \mathcal{F}(\mathcal{H}(\textbf{x}_{t-1}),\textbf{x}_{t-\tau:t-1})
\end{equation}

Similarly, the measurement noise is found using the equation:

\begin{equation} \label{eq:mn}
    \textbf{r}_t = \textbf{x}_t - \mathcal{H}^{-1}(\mathcal{H}(\textbf{x}_t))
\end{equation}

Using $\textbf{q}_t$ and $\textbf{r}_t$, the process covariance matrix $Q$ and measurement noise covariance matrix $R$ can be found, which we use later in our state estimation equations.

For training the neural network, we followed the guidelines provided in \cite{feng2021time}. To guide the hyperparamter tuning of the model, we used the uniform negative sampling algorithm to generate anomalous values. Next, a randomized grid search algorithm was used to fine-tune the hyperparameters and the model with the best anomaly detection performance. The hyperparameter summary is given in Table \ref{tab:hyperparams}.

\begin{table} \label{tab:hyperparams}
    \centering
    \caption{HyperParameters}
    \resizebox{0.5\textwidth}{!}{
        \begin{tabular}{ccc}
            \hline
            Hyperparameter & Dataset & Value or Range \\
            \hline
            \textit{$w_1$}  & ALL & 0.45 \\
            \textit{$w_2$} & ALL & 0.45 \\
            \textit{$w_3$} & ALL & 0.45 \\

            \textit{$\tau$} & PUMP & 5 \\
            \textit{$\tau$} & WADI & 12 \\
            \textit{$\tau$} & SWAT & 12 \\
            \textit{$\tau$} & HTTP & 12 \\
            \textit{$\tau$} & ASD & 12 \\
            NUM. Dimensions For \textit{$z_t$} & PUMP & [1, 120] \\
            NUM. Dimensions For \textit{$z_t$} & WADI & [1, 200] \\
            NUM. Dimensions For \textit{$z_t$} & SWAT & [1, 75] \\
            NUM. Dimensions For \textit{$z_t$} & HTTP & [1, 200] \\
            NUM. Dimensions For \textit{$z_t$} & ASD & [1, 200] \\
            NUM. Hidden Layers for $\mathcal{H}$ & ALL & [1, 3] \\
            NUM. Dense Layers for $\mathcal{F}$ & ALL & [1, 3] \\
            NUM. LSTM Layers for $\mathcal{F}$ & ALL & [1, 3] \\
            NUM Dimensions in Hidden Layers & ALL & [32, 256] \\
            Training Epochs & ALL & 100 \\
            \hline
        \end{tabular}
    }
\end{table}

\subsection{Anomaly Detection using Bayesian State-Space Module (BSSM)} \label{ssm}

In this section, we explain the implementation of the BSSM and designing of the state-space method. The BSSM is applied to the data set that contains both normal and anomalous data (denoted $X^{(a)}$). The state-space algorithm calculates the posterior probability $p(\textbf{z}_t|\textbf{x}_{t-\tau:t-1})$ and determines a value $\textbf{x}_t$ to be an anomaly if it is not part of the posterior distribution. Using the state transition function and measurement function from the NNM, the state-space equations are:

\begin{equation} \label{eq:3}
    \textbf{z}_t = \mathcal{F}(\textbf{z}_{t-1}, \textbf{x}_{t-\tau:t-1}) + \textbf{q}_t
\end{equation}

\begin{equation} \label{eq:4}
    \textbf{x}_t = \mathcal{H}^{-1}(\textbf{z}_{t}) + \textbf{r}_t
\end{equation}

It is assumed that both $\textbf{q}_t$ and $\textbf{r}_t$ are sampled from a Gaussian distribution. Hence, $\textbf{q}_t \backsim \mathcal{N}(\textbf{z}_0, Q)$ and $\textbf{r}_t \backsim \mathcal{N}(0, R)$ where $Q$ is the process noise covariance matrix and $R$ is the measurement noise covariance matrix, $\hat{\textbf{z}}_0=\mathcal{H}(\textbf{x}_0)$ and $P_0=\alpha\mathcal{I}$, where $I$ is the identity matrix. $\alpha$ is a factor for scaling the covariance matrix.

\subsubsection{Ensemble Kalman Filter Design} \label{EnKF Design}

In this section we explain the implementation detail of EnKF using \eqref{eq:EnKF Predict} and \eqref{eq:EnKF Update}. The problem with implementing EnKF from the equations is that the filter did not converge. We implementing EnKF using equations obtained from NNM by sampling the sigma points from a normal distribution that is an estimation of the hidden state. We call our proposed algorithm Neural Network EnKF (NN-EnKF) (Algorithm \ref{alg:EnKF}).

\RestyleAlgo{ruled}

\begin{algorithm} \label{alg:EnKF}
    \caption{Neural Network Ensemble Kalman Filter (NN-EnKF)}
    \KwData{Input multivariate time series data, $X \epsilon \mathbb{R}^{M \times T}$, M sensors and T time values}
    \KwResult{Estimated states, $\hat{\bm{\mu}} \epsilon  \mathbb{R}^{1 \times M}, \hat{P} \epsilon  \mathbb{R}^{M \times M}$}
    \textbf{Initialization:} $\bm{\chi} \backsim \mathcal{N}(\hat{\textbf{z}}_0, P_0)$\;
     \For{$i = 1:T$}{
            \tcc{Predict using \eqref{eq:EnKF Predict}}
            $\bm{\chi}_h = \mathcal{F}(\bm{\chi_{t-1}}, \textbf{x}_{t-\tau:t-1})$\; 
            \tcc{Update using \eqref{eq:EnKF Update}}
            $\bm{\chi}_f = \mathcal{H}^{-1}(\bm{\chi}_h)$\; 
            Find $P_{zz}, P_{xz} \& \mathcal{K}$\;
            Sample $\textbf{e}_r \backsim \mathcal{N}(0, R)$\;
            Update Sigma Points: $\bm{\chi}_t \leftarrow \bm{\chi}_f + \mathcal{K}[\textbf{x}_t - \bm{\chi}_h + \textbf{e}_r]$\; 
            Estimate state: $\hat{\bm{\mu}}_t \leftarrow \frac{1}{N} \sum_{i=1}^{N} \bm{\chi}_t$\;
            Estimate state covariance $\hat{P}_t \leftarrow P_{t-1} - \mathcal{K}P_{zz}\mathcal{K}^T$
        }
\end{algorithm}

Using \eqref{eq:EnKF Predict} and \eqref{eq:EnKF Update}, the updated equations are;

\begin{equation} \label{eq:enkf new}
    \begin{aligned}
        \bm{\chi}_h &= \mathcal{F}(\bm{\chi_{t-1}}, \textbf{x}_{t-\tau:t-1}) \\
        \bm{\chi}_f &= \mathcal{H}^{-1}(\bm{\chi}_h)
    \end{aligned}
\end{equation}

Where the sigma points $\bm{\chi}_t$ are used to represent the hidden state and is sampled from a normal distribution, $\bm{\chi}_t \backsim \mathcal{N}(\hat{z}_0, P_0), \bm{\chi}\epsilon\mathbb{R}^N$ where $N$ are the number of sigma points. $\bm{\chi}_h$ is the result after the state transition function is applied to $\bm{\chi}_t$ and $\bm{\chi}_f$ are the sigma points transformed back from the hidden state.

For initialization, to enable the filter to converge, $\alpha$ was set at a high value (which was 100). It was observed that the state transition and process noise are set to $0$ as they did not seem to make any contributions to the accuracy of the state estimation. Instead, we sampled the error perturbation from a multivariate normal distribution with mean 0 and covariance $R$ - $ \textbf{e}_r \backsim \mathcal{N}(0, R),  \textbf{e}_r\epsilon\mathbb{R}^N$ where $N$ are the number of sigma points. 

Using $P_{xx}$ and $P_{xz}$, $\mathcal{K}$ can be calculated which is used to update the sigma points. The updated sigma points become the input for the next state. The mean and variance of the sigma points become the state estimate.

\subsubsection{Particle Filter Design} \label{PF Design}

The next novel algorithm we use in BSSM is the modified version of SIR-PF called Neural Network SIR PF (NN-SIR). The first step is choosing the importance density. We choose a Gaussian distribution as the importance density. The particles are sampled from a normal distribution, $x_t^i \backsim \mathcal{N}(\hat{\textbf{z}}_0, P_0), i=1:N_s$ where $N_s$ are number of particles and $P_0$ is found with a very small $\alpha$. The weights are initialized as $w_i = \frac{1}{N_s}$ where ${i=1,...N_s}$. The state mean and covariance can be estimated using the following equations:

\begin{equation} \label{eq:pf new}
    \begin{aligned}
        X_h &= \mathcal{F}(X_{t-1}, \textbf{x}_{t-\tau:t-1}) + \textbf{q}_t \\
        X_f &= \mathcal{H}^{-1}(X_h) + \textbf{r}_t \\
    \end{aligned}
\end{equation}

In the equation $X_{t-1}=x_{t-1}^i, i=1,...N_s$,  $X_h$ is the result after the state transition function is applied to $X_{t-1}$ and $X_f$ are the sigma points transformed back from the hidden state.

The state transition noise is sampled from a normal distribution defined by $\textbf{q}_t \backsim \mathcal{N}(\hat{\textbf{z}}_0, Q)$. The process noise is also sampled from a normal distribution defined by $\textbf{r}_t \backsim \mathcal{N}(0, R)$.

At each iteration, the weights are updated by finding the squared error distance between $X_h$ and $\textbf{x}_t$. To measure the distance, we use the radial basis kernel function (RBF). The RBF is defined as: 

\begin{equation} \label{eq:rbf}
    rbf(\textbf{x}_t, X_f) = exp (\frac{||\textbf{x}_t - X_f||^2}{2\sigma^2})
\end{equation}

where $\sigma$ is a scaling number. The weights are then updated by $\mathcal{W}_u = \mathcal{W} * rbf(x_t, X_f)$ where $\mathcal{W}={w_t^i....w_t^{N_s}}$. The updated weights are like probabilities that determine how significant each particle is in relation to the data. The state mean and state covariance can then be determined from $X_h$ and $\mathcal{W}_u$. To avoid particle degeneracy, the particles are resampled using \textit{systematic} resampling as described in Section \ref{PF}. Systematic resampling gave the best performance for PF and we chose this resampling method. We set $N_t=0.1$. To further avoid degeneracy, $N_{rs}$ (we used $N_{rs}=1$) percent of the particles are resampled from the prior Gaussian distribution. The resampled particles are the input to the next state. The algorithm for the NN-SIR filter is given in Algorithm \ref{alg:PF}.

\begin{algorithm}  \label{alg:PF}
    \caption{Neural Network SIR Particle Filter (NN-SIR)}
    \KwData{Input multivariate time series data, $X \epsilon \mathbb{R}^{M \times T}$, M sensors and T time values}
    \KwResult{Estimated states, $\hat{\mu} \epsilon \mathbb{R}^{1 \times M}, \hat{P} \epsilon \mathbb{R}^{M \times M}$}
    \textbf{Initialization:} Sample $x_t^i \backsim \mathcal{N}(\hat{z}_0, P_0), i=1:N_s$ and assign weight $w_t^i=\frac{1}{N_s}$ for $i = 1:N_s$\;
    \For{$t=1:T$}{
        \For{$i=1:N_s$}{
        \tcc{propagate weights}
        $x_t^i = \mathcal{H}^{-1}((\mathcal{F}(x_{t-1}^i\textbf{x}_{t-\tau:t-1}) + \textbf{q}_t) + \textbf{r}_t$\; 
        normalize weights: $w_t^i = \frac{w_t^i}{\sum_{i=1}^{N_s}w_t^i}$\;
        }
        \tcc{Update Particles}
        \If{$\hat{N}_{eff} < N_T$} {$\{x_t^i, w_t^i\}_i^{N_s} = \texttt{RESAMPLE}(\{x_t^i, w_t^i\}_i^{N_s})$} 
        Resample $x_t^i \backsim \mathcal{N}(\hat{z}_0, P_0), i=1:N_s*N_{rs}$
        Estimate state: $\hat{\bm{\mu}}_t \leftarrow \frac{1}{N_s} \sum_{i=1}^{N} w_t^i x_t^i$ \;
        Estimate covariance: $\hat{P}_t \leftarrow \frac{1}{\sum_{i=1}^{N_s}w_t^i} \sum_{i=1}^{N_s} w_t^i (x_t^i - \hat{\bm{\mu}}_t)$ 
    }
\end{algorithm} 

\subsubsection{Anomaly Detection} \label{ad}

Let the predicted state mean and covariance be represented as $\hat{\bm{\mu}}_t$ and $\hat{P}_t$. The anomaly score can be found as the distance between the actual state and the predicted mean and covariance. To the find the distance, we use the Mahalanobis distance \cite{labbe2014kalman}. The Mahalanobis distance is defined as:

\begin{equation} \label{eq:maha}
    m = \sqrt{(\textbf{x}_t - \hat{\bm{\mu}}_t)^T \hat{P}_t^{-1} (\textbf{x}_t - \hat{\bm{\mu}}_t)}
\end{equation}

A larger $m$, indicates a higher likelihood for $\textbf{x}_t$ to be an anomaly. The anomaly threshold can be adaptive based on previous values or a hard threshold can be set or using the algorithm defined in \cite{du2021gan}. A value is considered to be an anomaly if its Mahalanobis is above a specific threshold.

The complete algorithm for our model Bayesian State-Space Anomaly Detection (\texttt{BSSAD}) is given in Algorithm \ref{alg:bssad} 

\begin{algorithm}  \label{alg:bssad}
    \caption{Bayesian State-Space Anomaly Detection (\texttt{BSSAD})}
    \KwData{Input multivariate time series data, $X \epsilon \mathbb{R}^{M \times T}$, M sensors and T time values}
    \KwResult{Anomaly Scores ($m$)}
    \textbf{Initialization:}\\
    Split $X$ into normal only ($X^{(n)}$) and normal \& anomalous data ($X^{(a)}$)\;
    Split $X^{(n)}$ into $X^{(n)}_{train}$ and $X^{(n)}_{val}$\;
    Train neural network using $X^{(n)}_{train}$\;
    Estimate noise on $X^{(n)}_{val}$ using \eqref{eq:pn} and \eqref{eq:mn}\;
    Divide $X^{(a)}$ into window size $\tau$\;
    \For{$t=1:\frac{T}{\tau}$}{
        Estimate $\hat{\bm{\mu}}_t$ and $\hat{P}_t$ for $X^{(a)}$ using Algorithm \ref{alg:EnKF} or \ref{alg:PF}\;
        Calculate anomaly score ($m$) using \eqref{eq:maha}
    }
\end{algorithm}

\section{Experiments} \label{exp}

In this section, we describe our experiments, the experimental setup and the data used for the experiments.

\subsection{Datasets} \label{data}

For our analysis, we use five different data sets. All the continuous values were normalized. Any sensors that had discrete values were formatted using one-hot encoding. The data sets are:

\begin{table*} [ht] \label{tab:data}
    \centering
    \caption{Dataset Description}
        \begin{tabular}{cccccc}
            \hline
            Dataset & Train Samples & Test Samples & Validation/Train & Features & Percent Anomaly/Test \\
            \hline
            SWAT & 99360 & 89984 & 0.25/0.75 & 51 & 11.99 \\
            \hline
            WADI & 241921 & 15701 & 0.25/0.75 & 93 & 7.09 \\
            \hline
            PUMP & 76901 & 143401 & 0.25/0.75 & 44 & 10.05 \\
            \hline
            HTTP & 200001 & 367497 & 0.25/0.75 & 3 & 0.6 \\
            \hline
            ASD & 8640 & 4320 & 0.25/0.75 & 19 & 10.21 \\
            \hline
            
        \end{tabular}
\end{table*}

\begin{itemize} 
    \item \textbf{Secure Water Treatment} (\texttt{SWaT}) \cite{goh2016dataset}: This data set consists of data collected from a Secure Water Treatment system. The data is collected over 11 days of operation, of which 7 days of data show the normal operation and 4 days of data were collected under an attack scenario. The data has 51 sensors and actuators. In addition, 5 seconds of data were aggregated.
    \item \textbf{Water Distribution} (\texttt{WaDi}) \cite{ahmed2017wadi}: The data is collected over 16 days of operation, out of which 14 days of data showed normal operation and 2 days of data were collected under an attack scenario. The data has 123 sensors and actuators. In addition, 5 seconds of data were aggregated.
    \item \textbf{PUMP Data} (\texttt{PUMP}) \cite{feng2021time}: The data was collected from a water pump system from a small town. The data was sampled at every minute and collected over 5 months.
    \item \textbf{HTTP} (\texttt{HTTP}) \cite{liu2008isolation}: This data set is about HTTP service data. We are using a small subset of the data set which was obtained from \cite{cup1999http}. From the subset, we manually split the test and train based on the number of consecutive normal points of data. The first 200001 points were all normal values and allocated towards the training set. The remaining 367497 points consist of mixed normal and abnormal values that were allocated towards the testing set.
    \item \textbf{Application Server Dataset} (\texttt{ASD}) \cite{su2019robust}: This public dataset contains data collected from an unspecified internet company. The data set consists of 12 files. For the training set, we used 8640 training samples per file. For the testing set, we used 4320 samples per file.
\end{itemize}

The details of the data set are shown in Table \ref{tab:data}.

\subsection{Analysis Metrics} \label{metrics}

A desirable method for measuring accuracy of anomaly detection is to use point-adjusted metrics \cite{xu2018unsupervised}. In point-adjusted metrics, a contiguous anomaly segment is considered as successfully detected if any point inside the segment is classified as an anomaly, Following the method described in \cite{feng2021time}, to measure the performance of our method, we enumerate all possible anomaly thresholds and search for the best possible point adjusted score.

For measuring the efficacy of the model, we use two metrics: \texttt{F1-score} (F1) \& \texttt{MCC}. F1 has been used in the past literature extensively as a good measure of accuracy of a model. 

The Matthew Correlation Coefficient (MCC) is a measurement metric for binary classification problems. It is considered as a better measurement metric than F1-score \cite{chicco2020advantages} as it considers all four parameters in the confusion matrix (false positive, true positive, false negative \& true negative). According to the literature F1 may be a too optimistic score for measuring accuracy in classification problems. Hence, we also use MCC as an additional metric for measuring the performance of our model. The MCC is defined as:

\begin{equation} \label{mcc}
    \resizebox{0.45\textwidth}{!}{$
        MCC = \frac{TP \cdot TN - FP \cdot FN}{\sqrt{(TP + FP) \cdot (TP + FN) \cdot (TN + FP) \cdot (TN + FN)}}
        $}
\end{equation}

where $TP$ are the true positives, $FP$ are the false positives and $FN$ are the false negative.

The MCC returns a value between $-1$ and $+1$. An MCC of $+1$ represents a perfect prediction, a 0 is no better than random prediction and $-1$ indicates total disagreement between predicted labels and actual labels.





\subsection{Benchmark Models} \label{bmodels}

\begin{table*} [ht]  \label{tab:results}
\centering
\caption{Comparison of Performance Metrics on Benchmark Datasets}
 \resizebox{0.95\textwidth}{!}{
  \begin{tabular}{cccccccccccccccc}
    \hline
    \multirow{2}{*}{Dataset} &
      \multicolumn{2}{c}{\texttt{SWaT}} &
      \multicolumn{2}{c}{\texttt{WaDi}} &
      \multicolumn{2}{c}{\texttt{PUMP}} &
      \multicolumn{2}{c}{\texttt{HTTP}} &
      \multicolumn{2}{c}{\texttt{ASD}} \\
    & F1 & MCC & F1 & MCC & F1 & MCC & F1 & MCC & F1 & MCC \\
    \hline
    IF & 0.9141 & 0.5081 & 0.8984 & 0.1086 & 0.8629 & 0.0252 & 0.9560 & 0.3391 & 0.8549 & -0.0247 \\
    \hline
    LSTM-AE & 0.8150 & 0.8084 & 0.7692 & 0.7550 & 0.8450 & 0.8301 & 0.9765 & 0.9764 & 0.6837 & 0.6550 \\
    \hline
    NSIBF & 0.9189 & 0.9107 & 0.9010 & 0.8928 & 0.9120 & 0.9208 & 0.9420 & 0.9420 & 0.8889 & 0.8785 \\
    \hline
    Interfusion & 0.8645 & 0.8572 & 0.8696 & 0.8598 & 0.7204 & 0.7213 & 0.6734 & 0.7100 & 0.9401 & 0.9335 \\
    \hline
    Anomaly Transformer & 0.7918 & 0.7732 & 0.9329 & 0.9299 & 0.8605 & 0.8528 & 0.4579 & 0.5405 & 0.9183 & 0.9089 \\
    \hline
    BeatGAN & 0.7549 & 0.7577 & 0.2409 & 0.1989 & 0.4186 & 0.3992 & 0.5888 & 0.6407 & 0.9183 & 0.9089 \\
    \hline
    FGANomaly & 0.9144 & 0.7206 & 0.5409 & 0.1237 & 0.9742 & 0.7378 & 0.0189 & 0.0486 & 0.8660 & 0.1267 \\
    \hline
    BSSAD-EnKF-10 & 0.8842 & 0.8679 & 0.7225 & 0.7007 & 0.9645 & 0.9606 & 0.8195 & 0.8197 & 0.7400 & 0.7052 \\
    BSSAD-EnKF-20 & 0.8997 & 0.8861 & 0.8061 & 0.7904 & \textbf{0.9838} & \textbf{0.9820} & 0.9714 & 0.9713 & 0.8636 & 0.8445 \\
    BSSAD-EnKF-50 & 0.9007 & 0.8895 & 0.7929 & 0.7888 & 0.9630 & 0.9594 & 0.6585 & 0.6946 & 0.8395 & 0.8231 \\
    \hline
    BSSAD-PF-500 & 0.9174 & 0.9088 & 0.9140 & 0.9084 & 0.8854 & 0.8806 & 0.9842 & 0.9842 & \textbf{0.9438} & \textbf{0.9359} \\
    BSSAD-PF-1000 & 0.9189 & 0.9107 & \textbf{0.9417} & \textbf{0.9386} & 0.8849 & 0.8799 & 0.9816 & 0.9816 & 0.9231 & 0.9125 \\
    BSSAD-PF-2000 & \textbf{0.9263} & \textbf{0.9538} & 0.9238 & 0.9203 & 0.8849 & 0.8799 & \textbf{0.9842} & \textbf{0.9844} & 0.9231 & 0.9125 \\
    \hline 
    \hline
  \end{tabular}
  }
\end{table*}

To demonstrate the benefits of our proposed methods, we compare our approach with the state-of-the-art methods, Specifically, we mainly focus on comparing with methods that have been released in the past few years:

\begin{itemize}
    \item Isolation Forest (IF) \cite{liu2008isolation}: It is a popular residual-error based framework, which is also part of the Python library. The method serves as a comparison between density-based and residual-error based anomaly detection algorithms.
    \item LSTM-AE \cite{feng2021time}: This is a type of residual-based anomaly detection method that combines LSTM and AE. This method will serve as a benchmark which shows the improvement of EnKF and PF over using only neural networks.
    \item Neural System Identification \& Bayesian Filtering (NSIBF) \cite{feng2021time}: Since our work is partially based on this paper and this work was shown to outperform all previous works, it seemed reasonable to use this paper as a benchmark for comparison. It also serves as a comparison of EnKF and PF over UKF.
    \item Interfusion \cite{li2021multivariate}: This work used Gibbs Sampler, a type of MCMC, for detecting anomalies. Since the focus of our work is on using the SMC methods, this was another work we chose to compare our model with.
    \item Anomaly-Transformer \cite{xu2022anomaly}: Method for anomaly detection using transformers. It was shown to outperform most of the other methods.
    \item BEATGAN \cite{du2021gan} \& FGANomaly \cite{zhou2019beatgan}: GAN based anomaly detection methods
\end{itemize}

\subsection{Results} \label{results}

All the experiments were run using a window size of 5 for the PUMP dataset and 12 for all the other datasets. For BSSAD we reported the best F1 and MCC scores, which were obtained by running the experiments repeatedly with different random seeds (we used 100 random seeds). We implemented our filters with a different numbers of sigma points and particles. The results of the experiments are shown in Table \ref{tab:results}. The results show that BSSAD performs best out of all the methods. PF has the highest F1 and MCC for both SWAT, WADI, HTTP and ASD data set compared to all three filters. EnKF shows the best performance for the PUMP data set. BSSAD is able to perform well on all data with small number of features and with small number of anomalies where most methods fail and perform poorly with very low F1 and MCC.

In addition, the MCC metric also shows that only measuring F1 might not always be an indication that a method is good at detecting anomalies. In some cases the F1 score was high but the MCC was very low. BSSAD is able to achieve high F1 and MCC showing that our method is very good at detecting anomalies.

The modularity of BSSAD shows that we can fine-tune our model based on the data. We could change the number of particles or sigma points to improve the accuracy. We could even swap out filters depending on the input data. For example, we could use EnKF for the PUMP data set and PF for WADI or SWAT data set. It should be noted that choosing EnKF or PF comes at a computation cost. There is a trade-off between the speed of detecting anomalies and the accuracy of detection. In domains where accuracy is more important, like in the medical domain, PF might be more suitable. But in domains where speed of detection is more important, like in detecting cyber-attacks, it might be more suitable to use EnKF or UKF.

In the future, we aim to explore in more detail the relationship between data and the type of filter. We will also explore other methods for NNM like hierarchical or variational autoencoders. Another avenue for future research would be to explore different importance densities or prior pdfs, other forms of PF, like particle Markov Chain Monte Carlos (PMCMC) \cite{chopin2020introduction}, a type of MCMC algorithm that incorporates particle filters to estimate the posterior distribution. Then samples are repeatedly obtained from the posterior to more accurately estimate the next state.

\section{Conclusion} \label{conc}

In this paper, we propose a novel and flexible model for anomaly detection in multivariate time series, called Bayesian State-Space Anomaly Detection (BSSAD). Our model combines neural networks and state-space algorithms to detect anomalies. The result of our analysis shows that BSSAD performs better than the state-of-the-art methods. Moreover, our method can be adapted for different data and domain by modifying state-space algorithms and neural networks to achieve better performance in detecting anomalies. In the future, we aim to further explore other state-space estimation methods and compare the differences in anomaly detection feasibility. We foresee the application of our model across multiple domains that use time-series data.

\bibliographystyle{IEEEtran}
\bibliography{anomaly_detection}


\end{document}